\documentclass[12pt]{article}

\expandafter\let\csname equation*\endcsname\relax
\expandafter\let\csname endequation*\endcsname\relax

\usepackage[numbers]{natbib}
\usepackage{hyperref}

\usepackage{amsmath,amssymb,amsfonts}
\usepackage{algorithmic}
\usepackage{graphicx}
\usepackage{textcomp}
\usepackage{xcolor}

\usepackage{tikz}
\usepackage{adjustbox}
\usepackage{natbib}

\DeclareMathOperator*{\argmax}{arg\,max}

\begin{document}

\title{Reinforcement Learning via Gaussian Processes with Neural Network Dual Kernels\\}

\author{Im\`{e}ne~R.~Goumiri\thanks{goumiri1@llnl.gov},\,
Benjamin~W.~Priest,\,
Michael~D.~Schneider 
\and 
\textit{\small Lawrence Livermore National Laboratory, Livermore, CA, 94551, USA.}
}

\date{\today\\
\vspace{0.1in}
{\footnotesize LLNL-JRNL-808440}
}

\maketitle

\begin{abstract}
While deep neural networks (DNNs) and Gaussian Processes (GPs) are both popularly utilized to solve problems in reinforcement learning, both approaches feature undesirable drawbacks for challenging problems. 
DNNs learn complex nonlinear embeddings, but do not naturally quantify uncertainty and are often data-inefficient to train.
GPs infer posterior distributions over functions, but popular kernels exhibit limited expressivity on complex and high-dimensional data.
Fortunately, recently discovered conjugate and neural tangent kernel functions encode the behavior of overparameterized neural networks in the kernel domain.
We demonstrate that these kernels can be efficiently applied to regression and reinforcement learning problems by analyzing a baseline case study.

We apply GPs with neural network dual kernels to solve reinforcement learning tasks for the first time.
We demonstrate, using the well-understood mountain-car problem, that GPs empowered with dual kernels perform at least as well as those using the conventional radial basis function kernel. 
We conjecture that  by inheriting the probabilistic rigor of GPs and the powerful embedding properties of DNNs, GPs using NN dual kernels will empower future reinforcement learning models on difficult domains.
\end{abstract}

\section{Introduction} \label{sec:intro}

The traditional approach in optimal control posits a controller with a suite of control signals able to affect a known dynamical system. 
The problem is to devise a policy for scheduling control signals in order to achieve some given objective.
As there is no uncertainty in the model, finding such a policy becomes an optimization problem.

However, many applications involve decision-making challenges where data are limited and the generative models are complex and partially or completely unknown. 
As such, the reinforcement learning (RL) branch of machine learning arose to develop models for an agent or agents acting on an initially unknown environment.
RL algorithms learn a policy to guide agent actions in order to achieve some high-level goal by acting on its environment and using the response to model its dynamics.

Although RL and optimal control are related, these research fields are traditionally separate. 
Ultimately, both are concerned with sequential decision making to minimize an expected long-term cost. 
The dynamical system, controller, and control signals of optimal control roughly map onto the environment, agent(s), and actions of RL.

Many RL algorithms \cite{Sutton18, Busoniu17, Lewis12, Vamvoudakis17, Zhu18}  address a lack of dynamics knowledge by way of a reliance upon parametric adaptive elements or control policies whose number of parameters or features are fixed and predetermined. 
These parameters are usually then learned from data.
Deep neural networks (DNNs) are also used extensively in RL \cite{Mnih13, Mnih15, Zoph16}.
DNNs are attractive as they are known to have an excellent representative power \cite{Hinton12, Kalchbrenner13, Krizhevsky12}. 
However, tuning and training the parameters is a data-inefficient practice \cite{Nagabandi18}. 
Moreover, DNNs usually include no natural means of quantifying the uncertainty in their predictions \cite{Gal15}.
Thus trained models may overconfidently predict the unknown dynamics when the system operates outside of the observed domain.
Such overconfident prediction can lead to system instability, thereby making any controller stability results unachievable.

Nonparametric kernel methods such as Gaussian processes (GPs) \cite{rasmussen06} have also been applied to reinforcement learning tasks \cite{Murray02, Kuss04, Lawrence05, Ko07, Tuong08, Deisenroth09, Deisenroth13}.
GPs are popular in many areas of machine learning due to their flexibility, interpretability, and natural uncertainty quantification due to being Bayesian models.
However GP-related data-driven methods remain largely unexploited in optimal control. 

The choice of GP kernel function encodes our prior beliefs about the distribution of the function of interest and is a key part of modeling.
For example, Kuss and Rasmussen use the famous radial basis function (RBF) in a GP to solve the mountain-car problem, implying that the dynamics are believed to be very smooth \cite{Kuss04}.
However, GPs often struggle to learn the features of complex or high-dimensional data, worrying the researcher interested in extrapolating this approach to such domains.

Recent results have shown a duality between wide, random DNNs and GPs through the use of the conjugate kernels (CK)\cite{neal1996priors, cho2009kernel, daniely2016toward, lee2018deep, matthews2018gaussian} and neural tangent kernels (NTK) \cite{jacot2018neural, arora2019exact, yang2019fine, arora2019harnessing}.
These kernels capture, in a sense that will be made explicit in Section~\ref{sec:nngp}, the nonlinear feature embedding learned by the corresponding DNN architecture.
However, these kernels are at present mostly treated as an academic curiosity and have predominantly been applied to image classification problems \cite{arora2019exact}.


\section{Gaussian Processes, Neural Networks, and Dual Kernels} \label{sec:nngp}

We will briefly review GPs, DNNs, and the correspondence between GPs and infinitely wide Bayesian DNNs.
We will focus on the computation of the various models, and largely omit training details.

GPs are flexible, nonparametric Bayesian models that specify a \emph{prior distribution over a function} $f : \mathcal{X} \rightarrow \mathcal{Y}$ that can be updated by data $\mathcal{D} \subset \mathcal{X} \times \mathcal{Y}$.
Coarsely, a GP is a collection of random variables, any finite subset of which has a multivariate Gaussian distribution.
We say that $f \sim \mathcal{GP}(m(\cdot), k(\cdot, \cdot))$, where $m : \mathcal{X} \rightarrow \mathbb{R}$ is a mean function and $k : \mathcal{X} \times \mathcal{X} \rightarrow \mathbb{R}$ a positive definite covariance function with hyperparameters $\theta$.
In practice $m$ is usually assumed to be the zero function.
Let $\mathbf{f}$ be a vector of evaluations of $f$ at some finite $X = \{\mathbf{x}_1, \dots, \mathbf{x}_n \} \subset \mathcal{X}$.
Then $f \sim \mathcal{GP}(\mathbf{0}, k(\cdot, \cdot))$ implies that
\begin{equation} \label{eq:gp_prior}
\mathbf{f} = [f(\mathbf{x}_1), \dots, f(\mathbf{x}_n)]^\top \sim \mathcal{N}(\mathbf{0}, K_{\mathbf{ff}}).
\end{equation}
Here $K_{\mathbf{ff}}$ is an $n \times n$ matrix whose $(i,j)$th element is $k(\mathbf{x}_i, \mathbf{x}_j) = \text{cov}(f(\mathbf{x}_i), f(\mathbf{x}_j))$.
Such covariance matrices implicitly depend on $\theta$.

We will assume that we actually observe data $\mathbf{y} = \mathbf{f} + \boldsymbol{\epsilon}$, where $\boldsymbol{\epsilon}_i \sim \mathcal{N}(0, \sigma^2)$ is homoscedastic noise.
Denote by $\mathbf{f}_*$ the vector of (unknown) evaluations of $f$ at a finite $X_* = \{\mathbf{x}_1^*, \dots, \mathbf{x}_{n_*}^* \} \subset \mathcal{X}$, and for convenience define $Q_\mathbf{ff} = K_\mathbf{ff} + \sigma^2 I_n$.
The GP prior on $f$ implies that the joint distribution of $\mathbf{y}$ and $\mathbf{f}_*$ is in the same form as Equation~(\ref{eq:gp_prior}):
\begin{equation}
  \begin{bmatrix} \mathbf{y} \\ \mathbf{f}_*
  \end{bmatrix}
  = \mathcal{N} \left ( 0,
  \begin{bmatrix}
    Q_\mathbf{ff} & K_\mathbf{f*} \\
    K_\mathbf{*f} & K_{**}
  \end{bmatrix} 
      \right ).
\end{equation}
Here $K_\mathbf{*f} = K_\mathbf{f*}^\top$ is the cross-covariance matrix between $X_*$ and $X$, i.e. the $(i, j)$th element of $K_\mathbf{*f}$ is $k(\mathbf{x}_i^*, \mathbf{x}_j)$.
The predictive posterior distribution on $\mathbf{f}_*$ conditioned on $X$, $X_*$ and $\mathbf{y}$ is therefore given by
\begin{align} \label{eq:gp_posterior}
\begin{split}
  \mathbf{f}_* \mid X, X_*, \mathbf{y} &\sim \mathcal{N}(K_\mathbf{*f} Q_\mathbf{ff}^{-1}\mathbf{y}, K_\mathbf{**} - K_\mathbf{*f} Q_\mathbf{ff}^{-1} K_\mathbf{f*}). \\
\end{split}
\end{align}
The posterior mean of Equation~\ref{eq:gp_posterior} is often given as the prediction for $f$ on $X_*$ in GP machine learning.

The expressiveness of a GP is heavily dependent upon the choice of kernel function $k$.
Most common functions, for example the RBF kernel,
\begin{equation} \label{eq:rbf}
k_{\textrm{RBF}}(\mathbf{x}, \mathbf{x}^\prime) = \text{exp}\left ( -\frac{\|\mathbf{x} - \mathbf{x}^\prime\|_2^2}{\ell^2} \right ),
\end{equation}
exhibit limited expressiveness on complex data and impose sometimes inappropriate assumptions such as stationarity.
GPs also suffer from cubic scaling in the observation size, although a rich literature of approximations addresses this problem.
See \cite{liu2020gaussian} and \cite{heaton2019case} for reviews of GP scaling methods.  
Furthermore, the recent emergence of fast and scalable GP software is beginning to challenge the conventional wisdom concerning the intractability of GPs by exploiting hardware acceleration \cite{gardner2018gpytorch, wang2019exact}.

DNNs learn an embedding of inputs into a latent space by way of iteratively applying nonlinear transforms.
This embedding transforms highly-nonlinear data relationships into a linear feature space, allowing a final linear regression to produce predictions.
In contrast to GPs, DNNs are highly parametric, often utilizing more parameters than observations.
For this reason, DNNs often require large amounts of training data, and a vast literature has developed around heuristic training protocols.
While DNNs do not, in general, produce posterior distributions or exhibit robust uncertainty quantification, their popularity is due to good empirical performance on complex and high-dimensional data.

A DNN with $L$ layers and widths $\{n^\ell\}_{\ell=0}^L$ has parameters consisting of weight matrices $\{ W^\ell \in \mathbb{R}^{n^\ell \times n^{\ell - 1}} \}_{\ell = 1}^L$ and biases $\{ \mathbf{b}^\ell \in \mathbb{R}^{\ell} \}_{\ell =1}^L$.
We will assume the NTK parameterization and introduce hyperparameters $\sigma_w$ and $\sigma_b$, whose interpretation we will define in Section~\ref{sec:duals}.
The output of a DNN on input $\mathbf{x}$ is $\mathbf{h}^L(\mathbf{x})$, which is computed recursively as
\begin{equation}
\begin{split} \label{eq:mlp}
\mathbf{h}^1(\mathbf{x}) &= \frac{\sigma_w}{\sqrt{n^0}} W^1\mathbf{x} + \sigma_b b^1,\\
\mathbf{h}^\ell(\mathbf{x}) &= \frac{\sigma_w}{\sqrt{n^{\ell-1}}} W^\ell \phi \left (\mathbf{h}^{\ell - 1}(\mathbf{x}) \right) + \sigma_b b^\ell.
\end{split}
\end{equation}
Here $\phi(\cdot)$ is an element-wise scalar nonlinear activation function, such as the popular ReLU function: 
\begin{equation} \label{eq:relu}
\phi_{\text{ReLU}}(x) = \max \{0, x\}.
\end{equation}

\subsection{Dual Kernels} \label{sec:duals}

As we have noted, GPs and DNNs have different advantages and disadvantages.
Many attempts have been made to obtain ``the best of both worlds'' - the uncertainty quantification and interpretability of GPs along with the computational convenience and expressivity of DNNs.
Such efforts include Bayesian neural networks, which apply prior distributions to the weights of neural networks \cite{neal1996priors}, and applying GPs to feature vectors embedded by DNNs \cite{mairal2014convolutional}.
Interestingly, a direct correspondence between GPs and Bayesian DNNs of any depth arises as the hidden layers become sufficiently wide. 
We will briefly motivate this correspondence, its history and applications.

Initializing all of the parameters in a DNN as $W^\ell_{i,j} \sim \mathcal{N} \left ( 0,\frac{\sigma^2_w}{n^{\ell -1}} \right )$ and $\mathbf{b}^\ell_{i} \sim \mathcal{N} \left ( 0,\sigma^2_b \right )$ for $i \in \left [ n^\ell \right ]$ and $j \in \left [n^{\ell - 1} \right ]$ (i.e. Glorot initialization \cite{lee2018deep}) is common in practice. 
Note that this is the equivalent of initializing all of the parameters in Eq. (\ref{eq:mlp}) as i.i.d.~$\mathcal{N}(0, 1)$.
In the study of highly overparameterized (wide) models over the last several decades, investigators made two unexpected observations.
\begin{enumerate} 
\item Random initialization followed by training only the final linear layer often produces high-quality predictions.
\item Training overparameterized models tends to produce weights that differ only slightly from initialization.  
\end{enumerate}

The correspondence between infinitely wide single hidden layer neural networks with i.i.d.~Gaussian weights and biases and analytic Gaussian Processes kernels was first discovered as far back as the 1990s by Neal by application of the Central Limit Theorem \cite{neal1996priors}.
Recently, others have extended Neal's result to infinitely wide deep neural networks \cite{lee2018deep, matthews2018gaussian} and convolutional neural networks with infinitely many channels \cite{novak2019bayesian, garriga2018deep}.
Arora et al. improved these results by showing that the correspondence holds for \emph{finite} neural networks that are sufficiently wide \cite{arora2019exact} and showed empirical evidence that the kernel process behavior occurs at lower widths than theoretically guaranteed \cite{arora2019harnessing}.
This is to say that the aurthors found that DNN predictions tend to agree with those of their dual GP counterparts, even when the DNNs are much narrower than the known theorems for convergence require. 
The kernel corresponding to wide DNNs is referred to in the literature as the conjugate kernel (CK) \cite{daniely2016toward} or NNGP kernel \cite{lee2018deep}.

In order to express the DNN transform Equation~(\ref{eq:mlp}) in kernel language, we must obtain a dual form of the nonlinearity $\phi$ that nonlinearly embeds a kernel matrix $K$ in another Hilbert space \cite{cho2009kernel, daniely2016toward}.
For nonlinearity $\phi$ and kernel matrix $K$ the dual form is known to be
\begin{equation} \label{eq:dual_activation}
V_\phi(K)(\mathbf{x}, \mathbf{x^\prime})
= \underset{f  \sim \mathcal{N}(\mathbf{0}, K)}{\mathbb{E}} \phi(f(\mathbf{x})) \phi(f(\mathbf{x}^\prime)).
\end{equation}
Using this dual transform and the notation of Eq. (\ref{eq:mlp}) and following the formulation of \cite{yang2019fine}, we can express the CK recursively as
\begin{align} \label{eq:ck}
\begin{split}
\Sigma^1(\mathbf{x}, \mathbf{x}^\prime) 
&= \frac{\sigma^2_w}{n^0} \langle \mathbf{x}, \mathbf{x}^\prime \rangle + \sigma^2_b \\
\Sigma^\ell(\mathbf{x}, \mathbf{x}^\prime) 
&= \sigma^2_w V_\phi(\Sigma^{\ell -1}) (\mathbf{x}, \mathbf{x}^\prime) + \sigma^2_b.
\end{split}
\end{align}
The last layer kernel $\Sigma^L$ is the conjugate kernel for the network. 
This kernel corresponds exactly to that of the linear model resulting from randomly initializing all weights and training the last layer. 

If the CK lends mathematical rigor to observation 1) above, the neural tangent kernel (NTK) does the same with observation 2). 
Intuitively, the NTK corresponds to a generalization of the CK where we train the whole model, rather than only the last layer.
The NTK emerges from the observation that infinitely wide neural networks evolve as linear models under stochastic gradient descent \cite{jacot2018neural, lee2019wide} and has also been shown to generalize to convolutional and finite architectures \cite{arora2019exact}.
Evidence suggests that the NTK might be capable of learning more complex features than the CK \cite{yang2019fine}, and the NTK has recently been shown to deliver competitive predictions in an SVM on small data learning benchmarks \cite{arora2019harnessing}.
We will omit the derivation, which is somewhat involved, and instead recite the form of the NTK $\Theta^L$ as given in \cite{yang2019fine}:
\begin{align} \label{eq:ntk}
\begin{split}
\Theta^1(\mathbf{x}, \mathbf{x}^\prime) 
&= \Sigma^1 (\mathbf{x}, \mathbf{x}^\prime) \\
\Theta^\ell(\mathbf{x}, \mathbf{x}^\prime) 
&= \Sigma^\ell(\mathbf{x}, \mathbf{x}^\prime) + \sigma^2_w \Theta^{\ell -1}(\mathbf{x}, \mathbf{x}^\prime) V_{\phi^\prime}(\Sigma^{\ell -1}) (\mathbf{x}, \mathbf{x}^\prime).
\end{split}
\end{align}

At first blush, the formulations of Eqs. (\ref{eq:ck}) and (\ref{eq:ntk}) are unhelpful, as computing Eq. (\ref{eq:dual_activation}) is intractable. 
Fortunately, closed-form solutions are known for several common activation functions \cite{cho2009kernel, daniely2016toward},  enabling efficient computation.
Throughout the rest of this document we will consider only networks utilizing $\phi_{\text{ReLU}}$, which is known to have analytic dual activations:
\begin{align} 
\label{eq:Vphi}
V_{\phi_{\text{ReLU}}} (K) (\mathbf{x}, \mathbf{x}^\prime)
&= \mbox{$\frac{\sqrt{K(\mathbf{x}, \mathbf{x}) K(\mathbf{x}^\prime, \mathbf{x}^\prime)}}{2 \pi} \left ( \sin c + ( \pi - c) \cos c) \right )$}  \\
\label{eq:Vphi'}
V_{\phi_{\text{ReLU}}^\prime}  (K) (\mathbf{x}, \mathbf{x}^\prime)
&= \frac{1}{2 \pi} ( \pi - c)  \\
\label{eq:c}
c 
&= \arccos \left ( \frac{K(\mathbf{x}, \mathbf{x}^\prime)}{\sqrt{K(\mathbf{x}, \mathbf{x}) K(\mathbf{x}^\prime, \mathbf{x}^\prime)}} \right ).
\end{align}

\subsection{A motivating example} \label{sec:toy}

We will illustrate the usage of the RBF kernel along with CK and NTK on a model of a simple machine. 
Consider the central example given in \cite{brynjarsdottir2014learning} of moving a weight up a slope.
We will assume that we are trying to learn a true process driven by the dynamics
\begin{equation} \label{eq:simple_dynamics}
\zeta(x \mid \theta, a) = \frac{\theta x}{1 - x / a}.
\end{equation}
Here $x$ is a control parameter modeling the amount of force exerted on the system, while $\theta$ and $a$ are unknown.
In terms of the model, the numerator of Eq.~(\ref{eq:simple_dynamics}) corresponds to the ideal efficiency of the machine, while the denominator corresponds to inefficiency (such as loss due to friction).

Say that we wish to model Eq. (\ref{eq:simple_dynamics}) using a GP, and that we have observed a vector of responses $\mathbf{y}$ at 11 locations $\mathbf{x}$  evenly-spaced in $[0.1, 4]$.
Then we believe that for each $i \in [11]$,
\begin{equation} \label{eq:simple_model}
\mathbf{y}_i = f(\mathbf{x}_i) + \epsilon_i
\end{equation}
where $f \sim \mathcal{GP}(\mathbf{0}, k(\cdot, \cdot))$ for some kernel function $k$ and $\epsilon_i \sim \mathcal{N}(0, \sigma^2)$ is measurement noise.

We simulate the dynamics
\begin{equation} \label{eq:simple_simulation}
\mathbf{y}_i = \zeta(\mathbf{x}_i \mid 0.65, 10) + \epsilon_i
\end{equation}
for $\epsilon_i \sim \mathcal{N}(0, 0.1^2)$ and fit GP models for each of $k_{\text{RBF}}$, $k_{\text{CK}}$, and $k_{\text{NTK}}$ as defined above.
We use Eq.~(\ref{eq:gp_posterior}) to learn the posterior distributions of each GP over a set $\mathbf{x_*}$ uniformly spaced in $[0.2, 9]$ and fit the hyperparameters of each kernel by way of a simple grid search using the loglikelihood.
See \cite{rasmussen06} Chapter 2 for a comprehensive review of GP regression.

\begin{figure}
  \includegraphics[width=\linewidth]{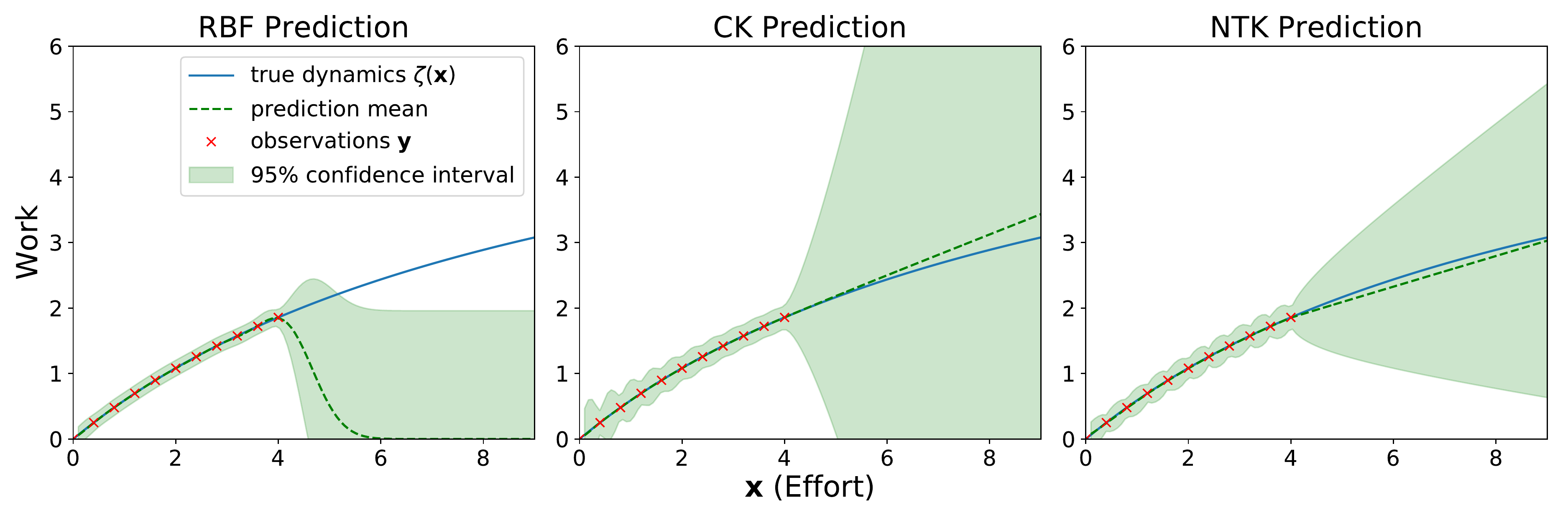}
  \caption{Predictive distributions of the RBF, CK and NTK trained on observations derived from the dynamics $\zeta(\cdot)$.}
  \label{fig:simple_machine}
\end{figure}

Figure~\ref{fig:simple_machine} plots the means of the resulting distributions and their 95\% confidence intervals, along with the true dynamics in blue and the observations in red.
Note that the RBF GP returns to the prior mean 0 when extrapolating far from the observed data. 
This is expected of stationary kernels, as inputs that are far apart are assumed to have low correlation. 
As given in Eqs.~(\ref{eq:ck}) and (\ref{eq:ntk}), both the CK and NTK kernels are functions of $\langle \mathbf{x}, \mathbf{x}^\prime \rangle$, $\| \mathbf{x} \|$, and $\| \mathbf{x}^\prime \|$.
Thus, they are \emph{nonstationary} on $\mathbb{R}^{n^0}$.
It is worth noting that most extant GP applications of CK and NTK use image data that has been normalized to the unit hypersphere \cite{lee2018deep, matthews2018gaussian, jacot2018neural, arora2019exact}.
In this case, CK and NTK are functions of the angle between the unit vectors $\mathbf{x}$ and $\mathbf{x}^\prime$, which maps one-to-one with $\| \mathbf{x} - \mathbf{x}^\prime \|$.
Consequently, in the aforementioned applications CK and NTK are isotropic. 
We do not perform normalization nor do we embed our data in a higher dimensional hypersphere in this work, meaning that in all cases the CK and NTK kernels are nonstationary. 

The fact that the posterior means of CK and NTK trend closer to the true dynamics far from the training data does not imply that these kernels are somehow ``better'' than RBF, but rather that their implicit assumptions about how the data is organized happen to center relatively well on this example.
More careful accounting of model discrepancy, such as that demonstrated in \cite{brynjarsdottir2014learning}, can produce much better extrapolation.
Note that in all cases, however, the confidence interval grows dramatically as we move further from the observed data. 
This behavior indicates a low confidence in any projections in these data, which provides a good example of what is desired from uncertainty quantification.
The majority of practical problems involve high dimensional transformations that are much harder to visualize.
Thankfully, the posterior distributions still allow the investigator to detect where predictions are uncertain due to the presence of high variance.
The rest of this document concerns itself with such an application to reinforcement learning.

\section{The mountain-car reinforcement learning problem}
\subsection{Description}
\begin{figure}
	\includegraphics[width=\linewidth]{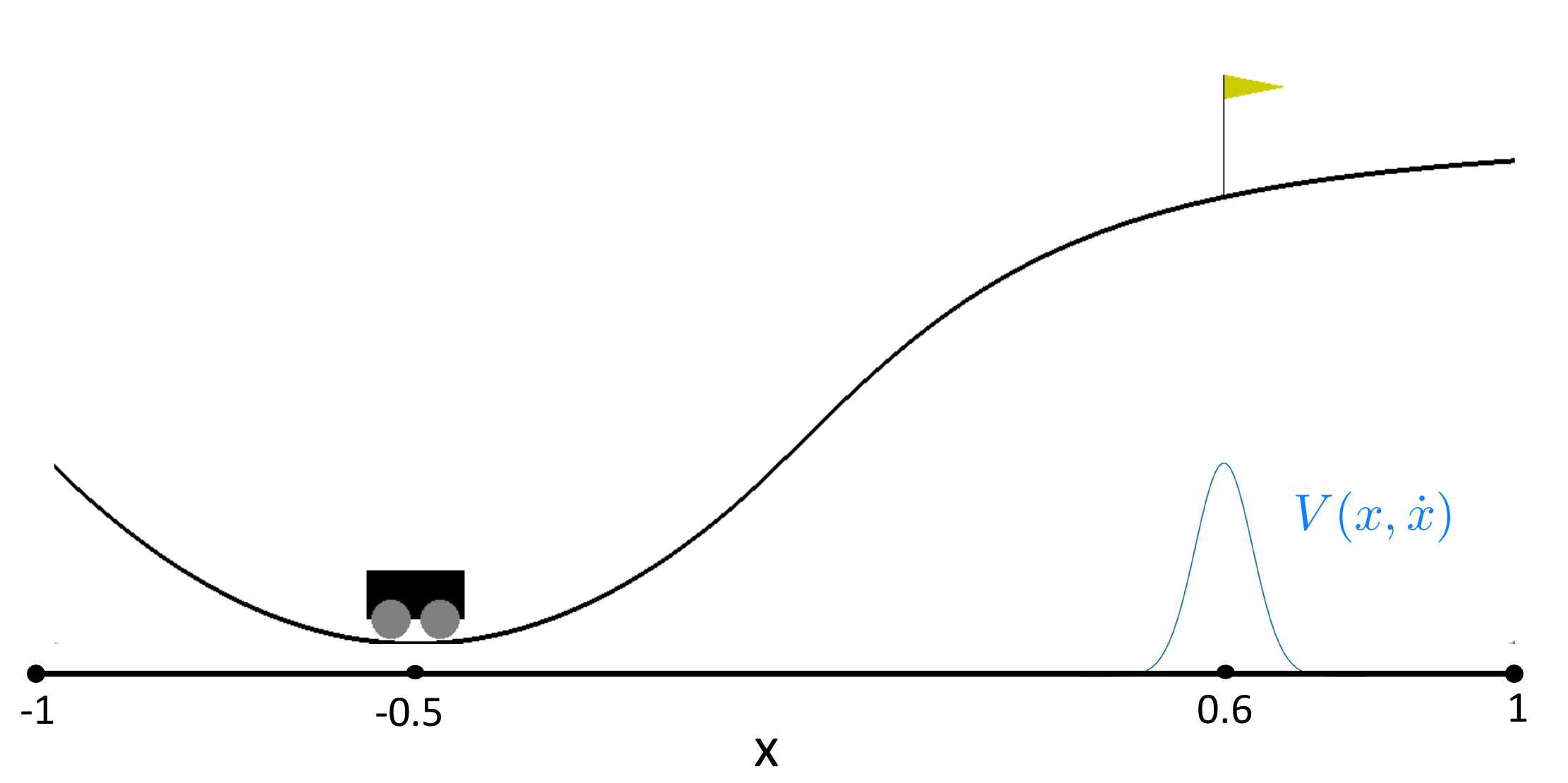}
	\caption{Illustration of the mountain car problem. The car is initially resting motionless at $x = -0.5$ and the goal is to bring it up and hold it in the region around the flag. }
	\label{mountaincar}
\end{figure}

The reinforcement learning problem studied in this paper is the mountain-car problem: a car drives along a mountain track and the objective is to drive to the top of the mountain. However gravity is stronger than the engine, and even at full thrust the car cannot accelerate up the steep slope. The only way to solve the problem is to first accelerate backwards, away from the goal, and then apply full thrust forwards, building up enough speed to carry over the steep slope even while slowing down the whole way. Thus, one must initially move away from the goal in order to reach it in the long run. This is a simple example of a task whose optimal solution is unintuitive: things must get worse before they can get better. The problem is fully described in \cite{Moore94, Sutton18} and is illustrated in Figure~\ref{mountaincar}.

The  mountain-car dynamical system has two continuous state variables, the position of the car $x$, and the velocity of the car $\dot{x}$. The state $s$ can be written as $s = \left(x,\dot{x} \right)$. The mountain surface is described by the altitude
\begin{equation}\label{eq:altitude}
H(x)=
\begin{cases}
 x^2 +x   & \text{if } x < 0,
\\
\frac{x}{\sqrt{1+5 x^2}}  & \text{if } x \geq 0.
\end{cases}
\end{equation}
The input that the driver can apply is the horizontal force $F$. Boundary conditions are imposed for each of the position, velocity and force of the car respectively as follows: 
\begin{align} \label{eq:bounds}
\begin{split}
-1 &\leq x \leq 1 \\
-2 &\leq \dot{x}\leq 2 \\ 
-4 &\leq F \leq 4.
\end{split}
\end{align}
The initial state $s_0 = [-0.5,0]^T$ indicates the car is at the unmoving at minimum altitude. 
This is the also the equilibrium of the dynamics.
The target reward $R$ is a multivariate gaussian PDF with mean $(x=0.6, \dot{x} =0)$ and covariance $\sigma^2 I_2$ with $\sigma = 0.05$.
$R$ is plotted in the top of Fig.~\ref{fig:rewardvalue}.

This choice of instantaneous reward function encodes into the model a desire to be as close as possible to the flag at position $x=6.0$ while remaining as stationary as possible.
The agent's goal is to find the optimal trajectory for the car to maneuver towards and remain near the flag, given the dynamics of the environment.
The core RL problem here is to find a {policy} for the decision maker (driver/car): a function  {$\pi$}  that specifies the action $F =\pi(s)$ that the decision maker will choose when in state $s$.

The standard family of algorithms to calculate this optimal policy constructs two arrays indexed by state: {policy $\pi$} and {value} ${V}$.
Upon completion of the algorithm, $\pi(s)$ specifies the action to be taken in state $s$, while $V(s)$ is the real-valued discounted sum of the rewards to be earned by following that solution from $s$.

This RL algorithm has two steps, (1) a value update and (2) a policy update, which are iterated across all the states until $\pi$ and $V$ converge. 
The Bellman equation is commonly used to update the value $V$:
\begin{equation} \label{eq:bellman}
{\displaystyle V(s) =\int P_{\pi (s)}(s,s')\left[R_{\pi (s)}(s,s')+\gamma V(s')\right] ds' }.
\end{equation}
Here $\gamma$  is the discount factor and satisfies $0 \leq \gamma \leq 1$, $P_{\pi (s)}$ is the transition probability of going from state $s$ to state $s'$ when applying action $\pi(s)$ and $R_{\pi (s)}$ is the corresponding immediate expected reward.
Given a computed value function $V$ for a given policy $\pi$, we can compute an implicitly optimized update policy $\pi^\prime$ as:
\begin{equation} \label{eq:update}
{\displaystyle \pi (s)=\operatorname {argmax} _{a}\left\{\int _{s'}P(s'\mid s,a)\left[R(s'\mid s,a)+\gamma V(s')\right] d s'\right\}}
\end{equation}
Section~\ref{sec:impl} explains the algorithm in detail.
The main idea is that we iterate the process of evaluating $V$ for a given policy $\pi$ over the continuous state space using Eq.~(\ref{eq:bellman}) and then recompute the policy using Eq.~(\ref{eq:update}).

\subsection{Algorithmic Implementation} \label{sec:impl}

Our algorithm is a generalization of the algorithm described in~\cite{Kuss04} which is able to accommodate the three different kernels described in the previous section while maintaining computational efficiency.
It proceeds by first initializing the dynamics of the model and value function, then iterating over updating the value and policy until convergence.
We model the dynamics using GPs.
In doing so, we explicitly solve the dynamics for a small number of observed position/velocity states, then train GPs to interpolate the state evolution of unobserved states.
Similarly, we use a separate GP to model the value function at a small number of position/velocity states, each with a small number uniform sample forces. 
We iterate over this GP, applying interpolation to update the learned policy which we in turn use to update the GP. We use the OpenAI Gym software~\cite{openai_gym} to model 
the dynamics environment.

\paragraph{\bf{Initialization of the dynamics}}
 
The first step is to train a GP to predict the dynamics of the system. The dynamical equation is
\begin{equation}\label{eq:dynamics}
  \frac{d}{dt}\begin{pmatrix} x \\ \dot{x} \\ F \end{pmatrix}
    = \begin{pmatrix} \dot{x} \\ F - G \cdot \sin(\arctan(H^\prime(x))) \\ 0 \end{pmatrix}.
\end{equation}
Here $G$ is the gravitational constant and $H^\prime$ is the derivative of the altitude given in Eq.~\eqref{eq:altitude} with respect to $x$. 
Given a state $s$, we integrate Eq.~(\ref{eq:dynamics}) forward in time over a span $\Delta t$ of 0.3\,s to obtain the corresponding \emph{next} state $s^\prime$. 
For training we take $N_d = 128$ random 3D states $s_i$ chosen uniformly in the domain defined by~\eqref{eq:bounds} and we compute their corresponding next states $s_i^\prime$.
We use these $s$ - $s^\prime$ pairs as observations to train two GPs, one for $x$ and one for $\dot{x}$.
We can then utilize Eq.~(\ref{eq:gp_posterior}) to interpolate the dynamics evolution at unobserved states.
We assume the hyperparameters of both CK and NTK to be distributed according to an inverse-Gamma distribution. 
We use a Monte Carlo Markov Chain technique to fit them by minimizing the mean square error when predicting the dynamics.
See \cite{rasmussen06} for a nuanced discussion of hyperparameter optimization.
Both kernels provide comparable accuracy for predicting the dynamics once trained and tuned.

\paragraph{\bf{Initialization of the value function}}

Next we must train a GP to predict the value function of any given state. 
The procedure is iterative so we use the reward $R$ as the initial value function. 
As with the dynamics, we take a certain number ($N_V = 512$) of random states in the 3D domain of $(s_j, F)$ uniformly from the domain~\eqref{eq:bounds} and associate them with their corresponding initial value ($\equiv R$) to provide training samples. 
Note that contrary to~\cite{rasmussen06}, we train that GP using the full 3D state $(x, \dot{x}, F)$ as input rather than omitting $F$ for reasons that will be apparent in the description of the iterations below. 
We tune the hyperparameters of the value GPs using the same MCMC procedure applied to the dynamics.

\paragraph{\bf{Iteration of the value and policy}}

Once all the GPs are trained and tuned, we can start iterating to update the value GP and the policy until the value function converges to a fixed point.
For each state $s_j$ in the dynamics training set, we generate a sequence of $N_F = 128$ states $s_k = (x_k, \dot{x}_k, F_k)$ where $\forall k$, $x_k = x_j$, $\dot{x}_k = \dot{x}_j$, and the actions $F_k$ are uniformly spaced and cover the entire $F$ domain. 
Then we use the dynamics GPs to predict their respective next states $s_k^\prime$ as the posterior means of Eq.~(\ref{eq:gp_posterior}).
$s_k^\prime$ then serves as input to the value GP to predict $V_k$, again as the posterior mean of Eq.~(\ref{eq:gp_posterior}). 
We can then compute $V_j^\text{max} = \max_k V_k$ and $k_\text{max} = \argmax_k V_k$ and deduce the policy $\pi(s_j) = F_{k_\text{max}}$. 
Finally we can update the value associated with each $s_j$ using $V_j \leftarrow R(s_j) + \gamma \cdot V_j^\text{max}$.

Once the value has been updated, we retrain the value GP. 
We repeat this procedure until the value stops evolving. 
Once it does we output the optimal policy $\pi$.

\subsection{Results}

We show that GPs using either both the CK and NTK as their kernel functions are suitable for solving the mountain car reinforcement learning problem.
The estimated dynamics are predicted with sufficient accuracy to enable the iterative evolution of the value function. 

\begin{figure}[!htb]
	\includegraphics[width=\linewidth]{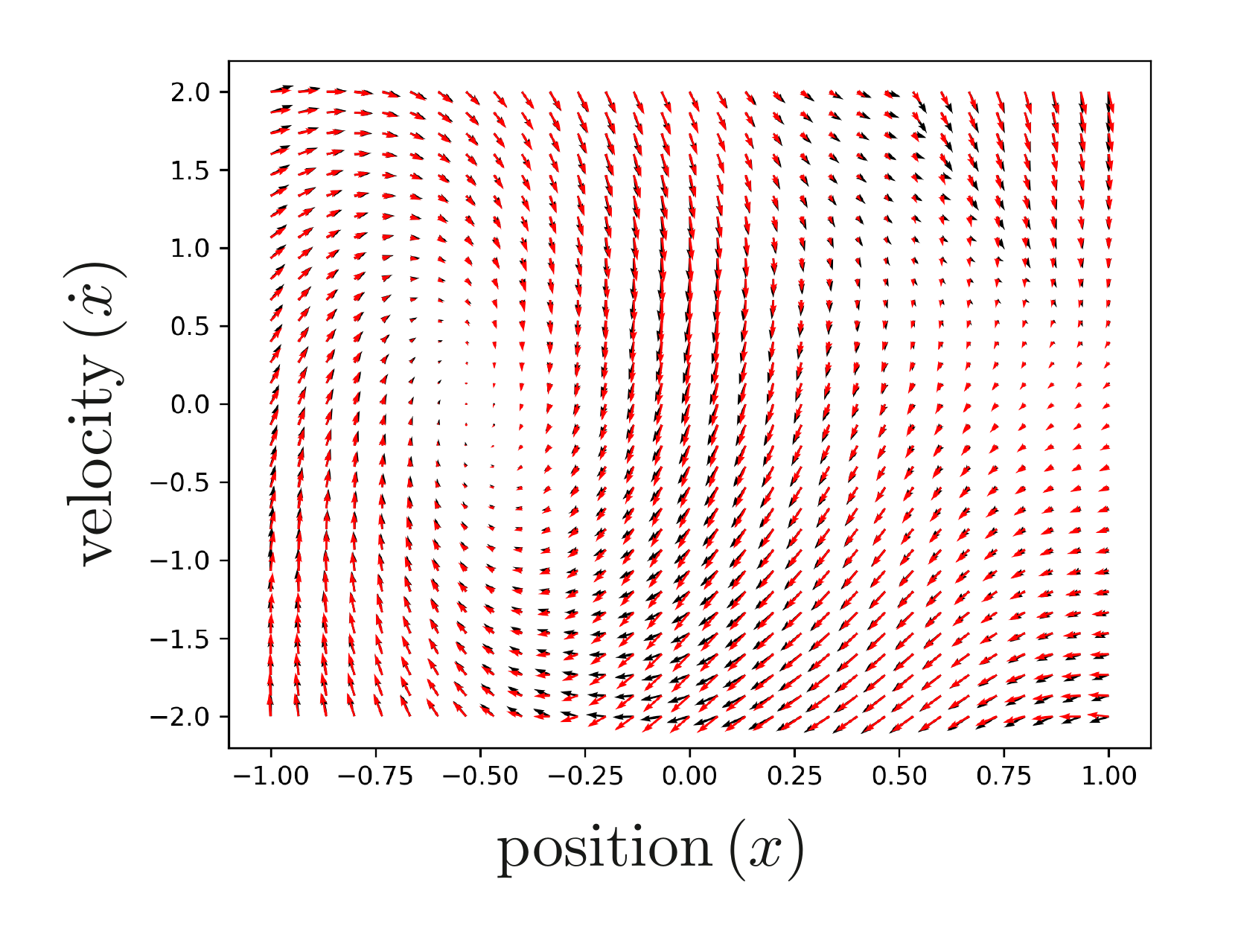}
	\caption{True (black) and predicted (red) dynamics as a function of $x$ and $\dot{x}$ (for $F = 0$). Each arrow represents a state $s$ (base of the arrow) and points in the direction of its \emph{next} state $s^\prime$ 0.3\,s in the future. The arrow lengths are scaled down so as not to overlap. The stable equilibrium at $(-0.5, 0)$ corresponds to the bottom of the valley. The target at (0.6, 0) is unstable requiring a sustained force $F > 0$ to maintain the car at the target. The discontinuity in the upper right of the phase plot is due to boundary conditions: hitting the boundary of the domain brings the velocity to zero.}
\label{fig:quiver}
\end{figure}

Figure~\ref{fig:quiver} shows the comparison between the dynamics derived from the physics (the truth) and the dynamics predicted using CK Gaussian process modeling. 
We can see that our dynamical model is very close to the reality. It captures the main features and equilibria of the dynamical system.

\begin{figure}[!htb]
	\centerline{
		\includegraphics[width=0.5\linewidth]{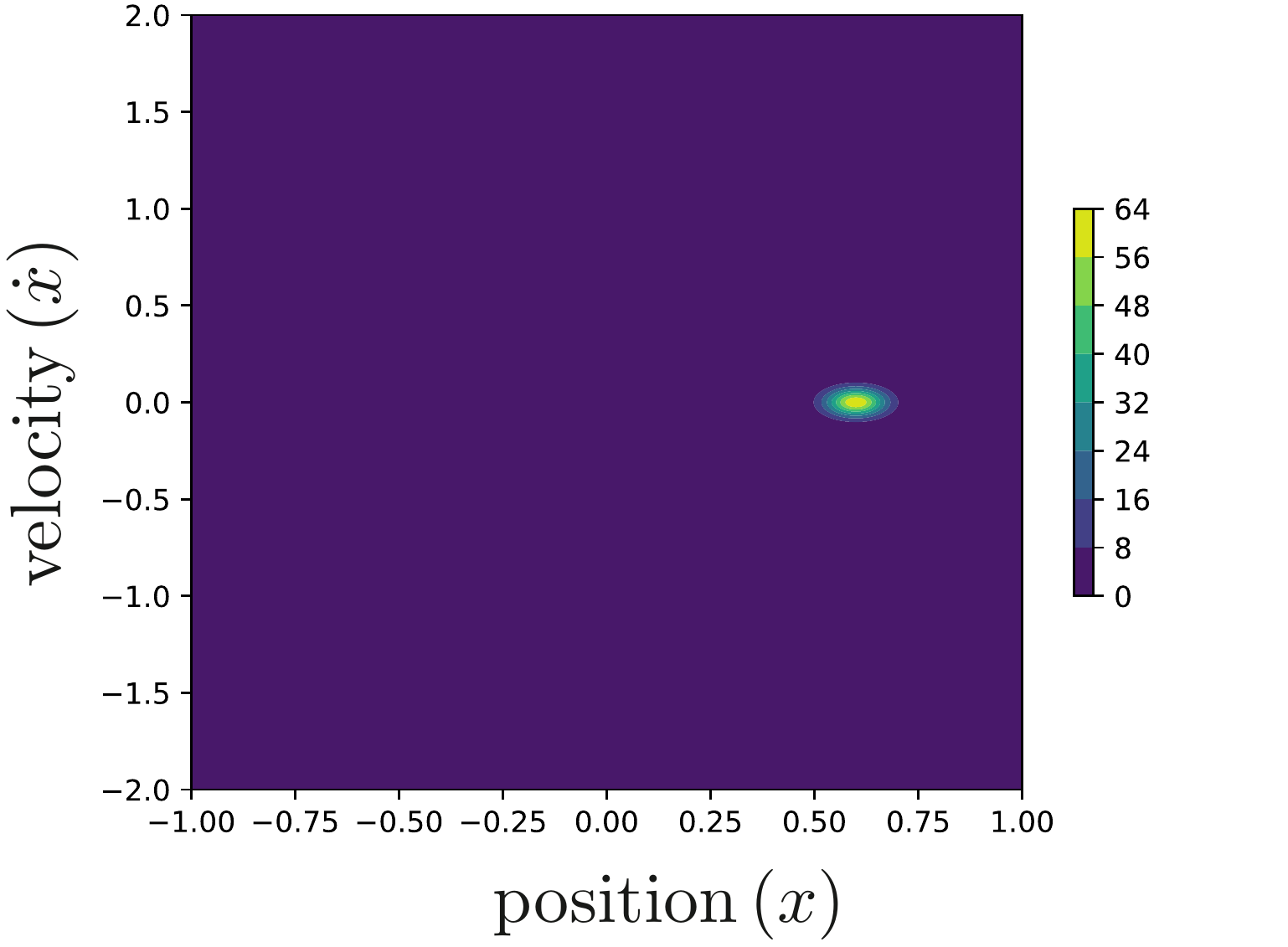}
	}
	\centerline{
		\includegraphics[width=0.5\linewidth]{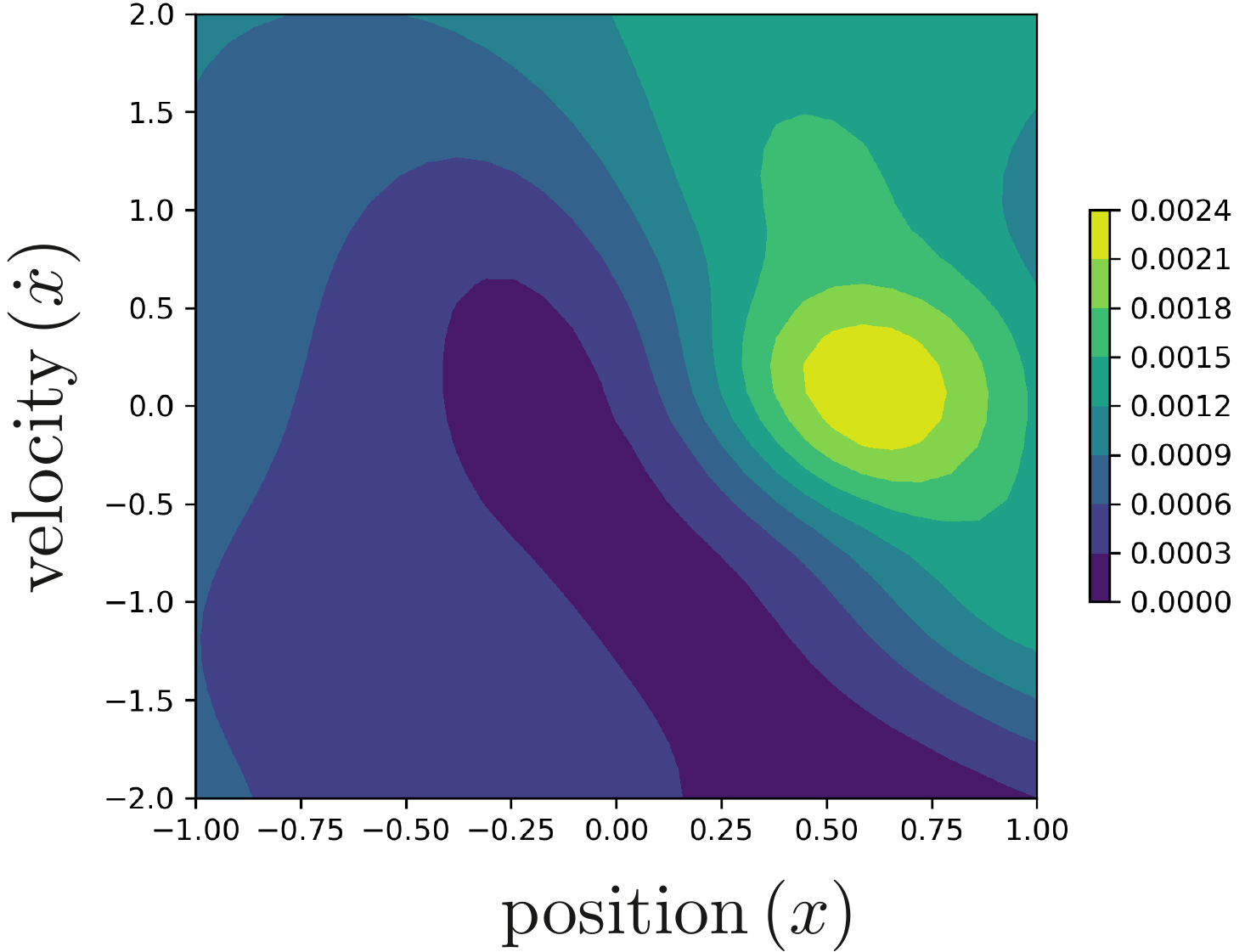}
		\includegraphics[width=0.5\linewidth]{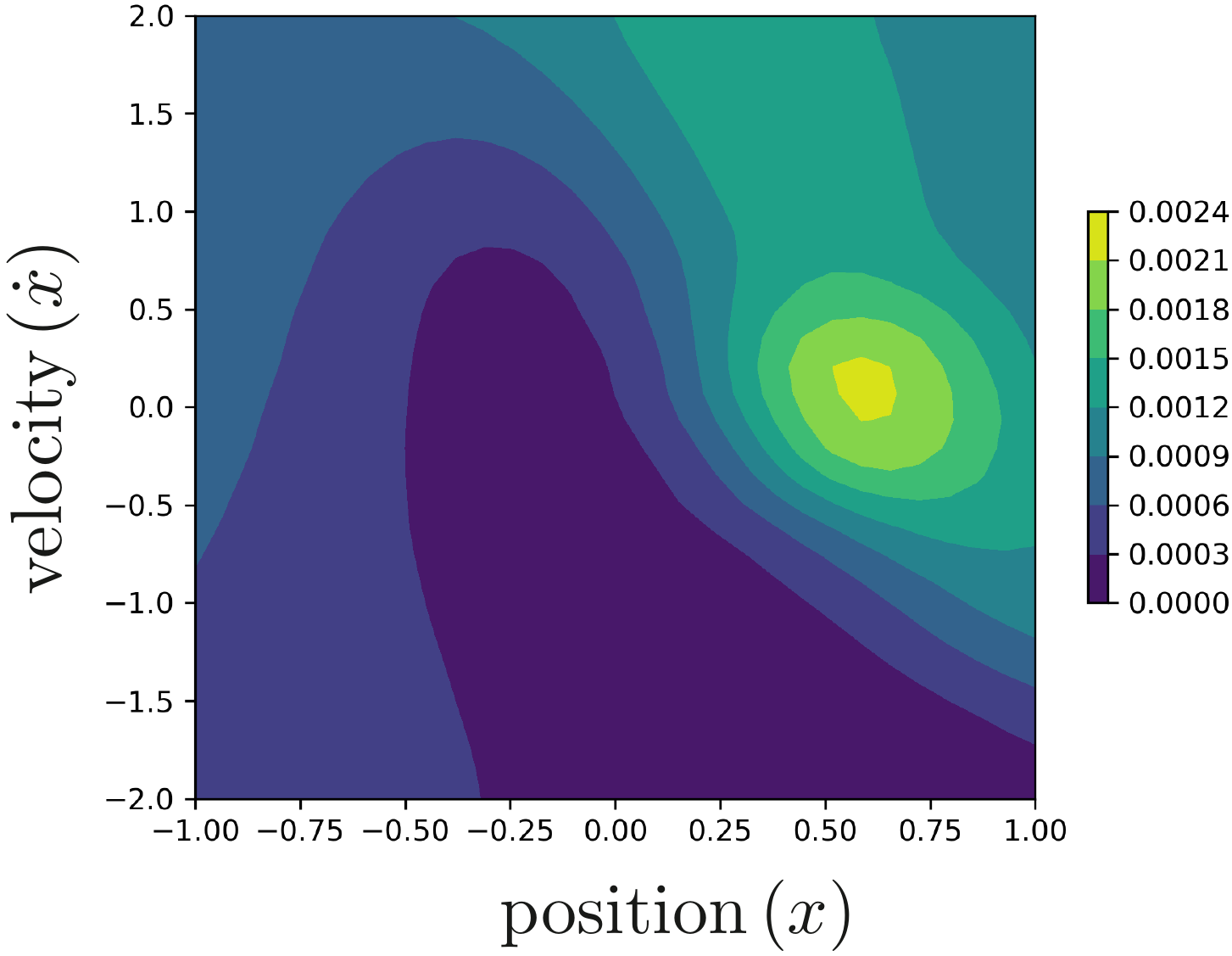}
	}
	\caption{Initial reward (top) and final value function for CK (bottom left) and NTK (bottom right) as a function of $x$ and $\dot{x}$ for $F=0$.}
	\label{fig:rewardvalue}
\end{figure}

Figure~\ref{fig:rewardvalue} shows the initial value (top) which is the instant reward, a Gaussian function centered around the target at $(x, \dot{x}) = (0.6, 0)$ with a small standard deviation of $0.05$, and the final value function for CK (middle) and NTK (bottom). 
For both kernels, the value function converges in six iterations. 
The value expands diagonally from the target to regions where the velocity is high enough to overcome the steep slope and finally curves back to reach the car's initial position from the left, leading to the non-trivial but correct policy that the car should start by going backward before speeding up the slope. 
The value function does not increase from zero in the central region of the phase space, which corresponds to the invalid policy of attempting to climb the slope of the mountain in the positive $x$-direction without sufficient momentum.

\begin{figure}[!htb]
	\centerline{
		\includegraphics[width=0.7\linewidth,trim={0.1cm 0.1cm 1.5cm 1.2cm},clip]{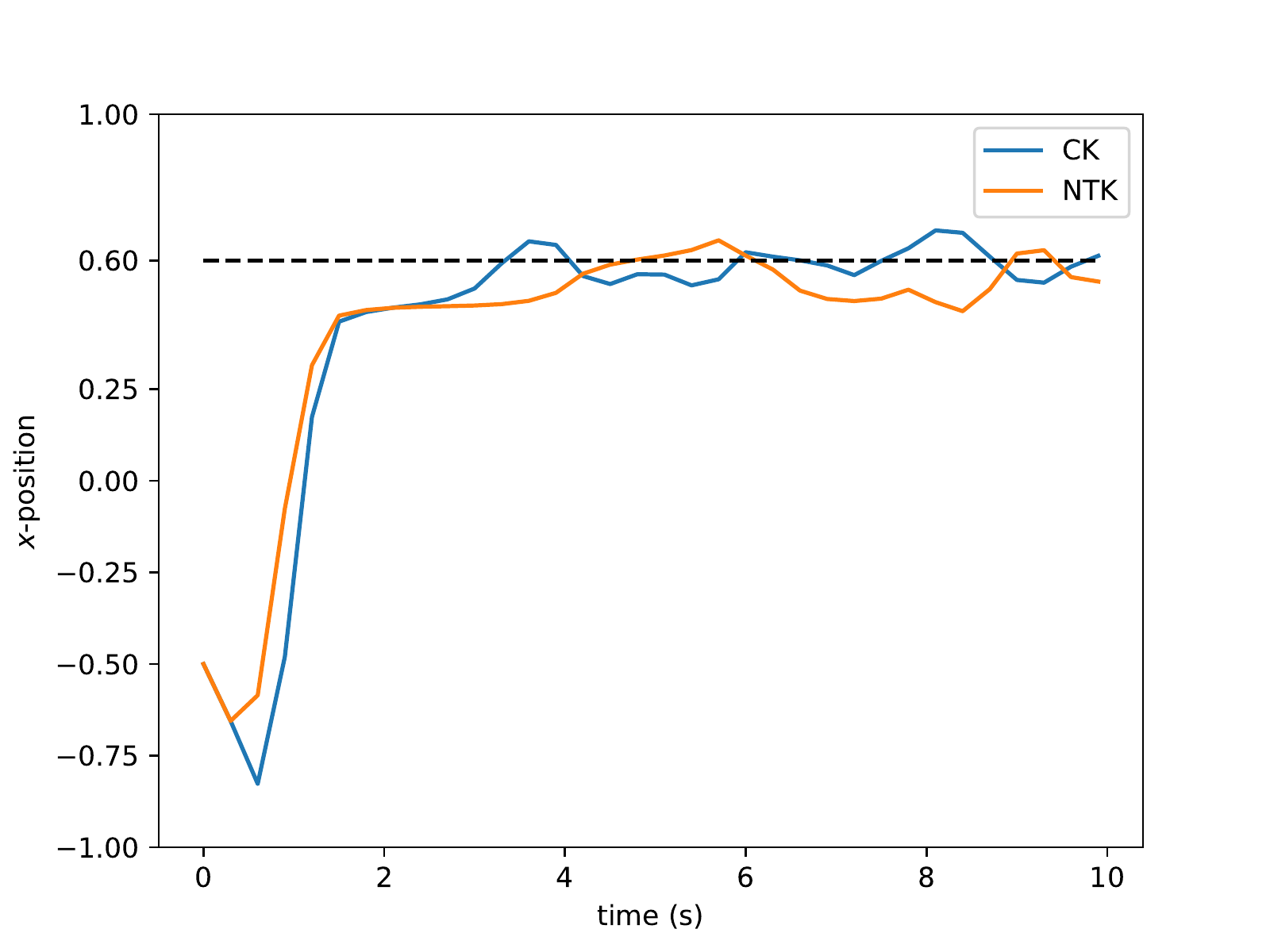}
	}
	\caption{Evolution of the $x$ position over time. The car goes backward initially then speeds up the hill.}
	\label{fig:trajectories}
\end{figure}

Figure~\ref{fig:trajectories} illustrates the learned optimal trajectories of the car along the $x$-axis over time. 
Again for both kernels, the car first moves backward then speeds up to quickly reach the target, where it stays indefinitely.
The oscillations of the car around the target location ($x=0.6 $) are the result of compounding errors in the GP predictions of the dynamics and the value. 
These oscillations can most likely be reduced by adding more training points, both for the dynamics and the value, but this comes at the cost of additional computation.

In our training of the GP representation of the value function, we see there are clear \emph{regions of interest} in the value function that change with each training iteration. 
This indicates that choosing training points uniformly in the entire domain and keeping the same points throughout iterations is suboptimal. 
It would be better to sample more points where the value is higher so as to achieve better resolution in this region.
In other words, we expect to see improved numerical performance by converting the value function into a probability density function for sampling training points followed by resampling the points after each policy iteration step to accommodate the changing value function. 
This should allow more accurate predictions without the performance cost of adding more training points.

\section{Perspectives and conclusion}

We have shown that GP kernels that are dual descriptions of neural networks are suitable for solving a simple reinforcement learning problem. 
The kernels we use here have been shown to perform well for GP classification tasks~\cite[e.g.,][]{lee2018deep}, but we believe our result is the first application of such kernels for GP regression in a non-trivial problem. 
We have also improved the GP model for the value function from those models presented in the literature~\cite[i.e.,][]{Kuss04} to increase the computational efficiency of the policy iteration step by decreasing the number of sample points in the combination of phase space and possible actions. 
We are able to achieve this performance improvement because of the improved expressivity of the GP regression in the combined sample space. 

While this simple mountain-car RL problem turns out to be easily soluble with GPs utilizing the classic RBF kernel, we have shown that neural network dual kernels deliver similar performance.
Furthermore, we expect that more challenging RL problems that have benefitted from neural networks for modeling the dynamics and the value function may also benefit in the future from the GP dual description of those networks~\cite[e.g.,][]{Mnih15}. 
In particular, RL problems relying on computer vision may benefit from application of the convolution version of the CK or NTK \cite{arora2019exact}.
Additionally, the ongoing development of kernels dual to arbitrary architectures opens up the possibility of taking advantage of recurrent neural network expressivity within the GP paradigm \cite{yang2019tensor}.
The GP dual to neural networks applied to RL thus offers promise of incorporating recent advances in deep RL with the probabilistic modeling features of GPs. 
Such applications also elude the grasp of more conventional GP models in the current literature due to the expressivity limitations of known kernels, especially on high-dimensional data.

We also have not fully exploited the value in utilizing dual GPs.
All of the predictions given throughout this document utilize only the posterior mean of Eq.~(\ref{eq:gp_posterior}) for prediction.
In this sense, we might as well have actually used overparameterized DNNs for prediction.
We expect that knowledge of the posterior variance will be greatly beneficial in more advanced RL problems where dynamics and value function propagation involves more uncertainty. 
We will incorporate applications of uncertainty quantification into future work.

\section*{Acknowledgments}
This work was performed under the auspices of the U.S. Department of Energy by Lawrence Livermore National Laboratory under Contract DE-AC52-07NA27344.
Funding for this work was provided by LLNL Laboratory Directed Research and Development grant 19-SI-004.

\bibliographystyle{plain}
\bibliography{../GPRL}

\begin{thebibliography}{10}

\bibitem{arora2019exact}
Sanjeev Arora, Simon~S Du, Wei Hu, Zhiyuan Li, Ruslan Salakhutdinov, and
  Ruosong Wang.
\newblock On exact computation with an infinitely wide neural net.
\newblock {\em arXiv preprint arXiv:1904.11955}, 2019.

\bibitem{arora2019harnessing}
Sanjeev Arora, Simon~S Du, Zhiyuan Li, Ruslan Salakhutdinov, Ruosong Wang, and
  Dingli Yu.
\newblock Harnessing the power of infinitely wide deep nets on small-data
  tasks.
\newblock {\em arXiv preprint arXiv:1910.01663}, 2019.

\bibitem{openai_gym}
Greg Brockman, Vicki Cheung, Ludwig Pettersson, Jonas Schneider, John Schulman,
  Jie Tang, and Wojciech Zaremba.
\newblock Openai gym, 2016.

\bibitem{brynjarsdottir2014learning}
Jenn{\`y} Brynjarsd{\'o}ttir and Anthony O'Hagan.
\newblock Learning about physical parameters: The importance of model
  discrepancy.
\newblock {\em Inverse problems}, 30(11):114007, 2014.

\bibitem{Busoniu17}
Lucian Busoniu, Robert Babuska, Bart De~Schutter, and Damien Ernst.
\newblock {\em Reinforcement learning and dynamic programming using function
  approximators}.
\newblock CRC press, 2017.

\bibitem{cho2009kernel}
Youngmin Cho and Lawrence~K Saul.
\newblock Kernel methods for deep learning.
\newblock In {\em Advances in neural information processing systems}, pages
  342--350, 2009.

\bibitem{daniely2016toward}
Amit Daniely, Roy Frostig, and Yoram Singer.
\newblock Toward deeper understanding of neural networks: The power of
  initialization and a dual view on expressivity.
\newblock In {\em Advances In Neural Information Processing Systems}, pages
  2253--2261, 2016.

\bibitem{Deisenroth13}
Marc~Peter Deisenroth, Dieter Fox, and Carl~Edward Rasmussen.
\newblock Gaussian processes for data-efficient learning in robotics and
  control.
\newblock {\em IEEE transactions on pattern analysis and machine intelligence},
  37(2):408--423, 2013.

\bibitem{Deisenroth09}
Marc~Peter Deisenroth, Carl~Edward Rasmussen, and Jan Peters.
\newblock Gaussian process dynamic programming.
\newblock {\em Neurocomputing}, 72(7-9):1508--1524, 2009.

\bibitem{Gal15}
Yarin Gal and Zoubin Ghahramani.
\newblock Dropout as a bayesian approximation: Insights and applications.
\newblock In {\em Deep Learning Workshop, ICML}, volume~1, page~2, 2015.

\bibitem{gardner2018gpytorch}
Jacob Gardner, Geoff Pleiss, Kilian~Q Weinberger, David Bindel, and Andrew~G
  Wilson.
\newblock Gpytorch: Blackbox matrix-matrix gaussian process inference with gpu
  acceleration.
\newblock In {\em Advances in Neural Information Processing Systems}, pages
  7576--7586, 2018.

\bibitem{garriga2018deep}
Adri{\`a} Garriga-Alonso, Carl~Edward Rasmussen, and Laurence Aitchison.
\newblock Deep convolutional networks as shallow gaussian processes.
\newblock {\em arXiv preprint arXiv:1808.05587}, 2018.

\bibitem{heaton2019case}
Matthew~J Heaton, Abhirup Datta, Andrew~O Finley, Reinhard Furrer, Joseph
  Guinness, Rajarshi Guhaniyogi, Florian Gerber, Robert~B Gramacy, Dorit
  Hammerling, Matthias Katzfuss, et~al.
\newblock A case study competition among methods for analyzing large spatial
  data.
\newblock {\em Journal of Agricultural, Biological and Environmental
  Statistics}, 24(3):398--425, 2019.

\bibitem{Hinton12}
Geoffrey Hinton, Li~Deng, Dong Yu, George Dahl, Abdel-rahman Mohamed, Navdeep
  Jaitly, Andrew Senior, Vincent Vanhoucke, Patrick Nguyen, Brian Kingsbury,
  et~al.
\newblock Deep neural networks for acoustic modeling in speech recognition.
\newblock {\em IEEE Signal processing magazine}, 29, 2012.

\bibitem{jacot2018neural}
Arthur Jacot, Franck Gabriel, and Cl{\'e}ment Hongler.
\newblock Neural tangent kernel: Convergence and generalization in neural
  networks.
\newblock In {\em Advances in neural information processing systems}, pages
  8571--8580, 2018.

\bibitem{Kalchbrenner13}
Nal Kalchbrenner and Phil Blunsom.
\newblock Recurrent continuous translation models.
\newblock In {\em Proceedings of the 2013 Conference on Empirical Methods in
  Natural Language Processing}, pages 1700--1709, 2013.

\bibitem{Ko07}
Jonathan Ko, Daniel~J Kleint, Dieter Fox, and Dirk Haehnelt.
\newblock Gp-ukf: Unscented kalman filters with gaussian process prediction and
  observation models.
\newblock In {\em 2007 IEEE/RSJ International Conference on Intelligent Robots
  and Systems}, pages 1901--1907. IEEE, 2007.

\bibitem{Krizhevsky12}
Alex Krizhevsky, Ilya Sutskever, and Geoffrey~E Hinton.
\newblock Imagenet classification with deep convolutional neural networks.
\newblock In {\em Advances in neural information processing systems}, pages
  1097--1105, 2012.

\bibitem{Kuss04}
Malte Kuss and Carl~E Rasmussen.
\newblock Gaussian processes in reinforcement learning.
\newblock In {\em Advances in neural information processing systems}, pages
  751--758, 2004.

\bibitem{Lawrence05}
Neil Lawrence.
\newblock Probabilistic non-linear principal component analysis with gaussian
  process latent variable models.
\newblock {\em Journal of machine learning research}, 6(Nov):1783--1816, 2005.

\bibitem{lee2018deep}
Jaehoon Lee, Yasaman Bahri, Roman Novak, Samuel~S Schoenholz, Jeffrey
  Pennington, and Jascha Sohl-Dickstein.
\newblock Deep neural networks as gaussian processes.
\newblock In {\em International Conference on Learning Representations}, 2018.

\bibitem{lee2019wide}
Jaehoon Lee, Lechao Xiao, Samuel~S Schoenholz, Yasaman Bahri, Jascha
  Sohl-Dickstein, and Jeffrey Pennington.
\newblock Wide neural networks of any depth evolve as linear models under
  gradient descent.
\newblock {\em arXiv preprint arXiv:1902.06720}, 2019.

\bibitem{Lewis12}
Frank~L Lewis, Draguna Vrabie, and Kyriakos~G Vamvoudakis.
\newblock Reinforcement learning and feedback control: Using natural decision
  methods to design optimal adaptive controllers.
\newblock {\em IEEE Control Systems Magazine}, 32(6):76--105, 2012.

\bibitem{liu2020gaussian}
Haitao Liu, Yew-Soon Ong, Xiaobo Shen, and Jianfei Cai.
\newblock When gaussian process meets big data: A review of scalable gps.
\newblock {\em IEEE Transactions on Neural Networks and Learning Systems},
  2020.

\bibitem{mairal2014convolutional}
Julien Mairal, Piotr Koniusz, Zaid Harchaoui, and Cordelia Schmid.
\newblock Convolutional kernel networks.
\newblock In {\em Advances in neural information processing systems}, pages
  2627--2635, 2014.

\bibitem{matthews2018gaussian}
Alexander G de~G Matthews, Mark Rowland, Jiri Hron, Richard~E Turner, and
  Zoubin Ghahramani.
\newblock Gaussian process behaviour in wide deep neural networks.
\newblock In {\em Internation Conference on Learning Representation}, 2018.

\bibitem{Mnih13}
Volodymyr Mnih, Koray Kavukcuoglu, David Silver, Alex Graves, Ioannis
  Antonoglou, Daan Wierstra, and Martin Riedmiller.
\newblock Playing atari with deep reinforcement learning.
\newblock {\em arXiv preprint arXiv:1312.5602}, 2013.

\bibitem{Mnih15}
Volodymyr Mnih, Koray Kavukcuoglu, David Silver, Andrei~A Rusu, Joel Veness,
  Marc~G Bellemare, Alex Graves, Martin Riedmiller, Andreas~K Fidjeland, Georg
  Ostrovski, et~al.
\newblock Human-level control through deep reinforcement learning.
\newblock {\em Nature}, 518(7540):529, 2015.

\bibitem{Moore94}
Andrew~W Moore.
\newblock The parti-game algorithm for variable resolution reinforcement
  learning in multidimensional state-spaces.
\newblock In {\em Advances in neural information processing systems}, pages
  711--718, 1994.

\bibitem{Murray02}
Roderick Murray-Smith and Daniel Sbarbaro.
\newblock Nonlinear adaptive control using nonparametric gaussian process prior
  models.
\newblock {\em IFAC Proceedings Volumes}, 35(1):325--330, 2002.

\bibitem{Nagabandi18}
Anusha Nagabandi, Gregory Kahn, Ronald~S Fearing, and Sergey Levine.
\newblock Neural network dynamics for model-based deep reinforcement learning
  with model-free fine-tuning.
\newblock In {\em 2018 IEEE International Conference on Robotics and Automation
  (ICRA)}, pages 7559--7566. IEEE, 2018.

\bibitem{neal1996priors}
Radford~M Neal.
\newblock Priors for infinite networks.
\newblock In {\em Bayesian Learning for Neural Networks}, pages 29--53.
  Springer, 1996.

\bibitem{Tuong08}
Duy Nguyen-Tuong and Jan Peters.
\newblock Local gaussian process regression for real-time model-based robot
  control.
\newblock In {\em 2008 IEEE/RSJ International Conference on Intelligent Robots
  and Systems}, pages 380--385. IEEE, 2008.

\bibitem{novak2019bayesian}
Roman Novak, Lechao Xiao, Jaehoon Lee, Yasaman Bahri, Greg Yang, Dan Abolafia,
  Jeffrey Pennington, and Jascha Sohl-dickstein.
\newblock Bayesian deep convolutional neural networks with many channels are
  gaussian processes.
\newblock In {\em International Conference on Learning Representation}, 2019.

\bibitem{Sutton18}
Richard~S Sutton and Andrew~G Barto.
\newblock {\em Reinforcement learning: An introduction}.
\newblock MIT press, 2018.

\bibitem{Vamvoudakis17}
Kyriakos~G Vamvoudakis, Hamidreza Modares, Bahare Kiumarsi, and Frank~L Lewis.
\newblock Game theory-based control system algorithms with real-time
  reinforcement learning: How to solve multiplayer games online.
\newblock {\em IEEE Control Systems Magazine}, 37(1):33--52, 2017.

\bibitem{wang2019exact}
Ke~Wang, Geoff Pleiss, Jacob Gardner, Stephen Tyree, Kilian~Q Weinberger, and
  Andrew~Gordon Wilson.
\newblock Exact gaussian processes on a million data points.
\newblock In {\em Advances in Neural Information Processing Systems}, pages
  14622--14632, 2019.

\bibitem{rasmussen06}
Christopher~KI Williams and Carl~Edward Rasmussen.
\newblock {\em Gaussian processes for machine learning}, volume~2.
\newblock MIT Press Cambridge, MA, 2006.

\bibitem{yang2019tensor}
Greg Yang.
\newblock Tensor programs i: Wide feedforward or recurrent neural networks of
  any architecture are gaussian processes.
\newblock {\em arXiv preprint arXiv:1910.12478}, 2019.

\bibitem{yang2019fine}
Greg Yang and Hadi Salman.
\newblock A fine-grained spectral perspective on neural networks.
\newblock {\em arXiv preprint arXiv:1907.10599}, 2019.

\bibitem{Zhu18}
Yuanheng Zhu and Dongbin Zhao.
\newblock Comprehensive comparison of online adp algorithms for continuous-time
  optimal control.
\newblock {\em Artificial Intelligence Review}, 49(4):531--547, 2018.

\bibitem{Zoph16}
Barret Zoph and Quoc~V. Le.
\newblock Neural architecture search with reinforcement learning.
\newblock {\em CoRR}, abs/1611.01578, 2016.

\end{thebibliography}

\end{document}